%% file: main.tex
\documentclass[10pt,twocolumn,letterpaper]{article}

\usepackage[pagenumbers]{cvpr} 

\input{preamble}

\definecolor{cvprblue}{rgb}{0.21,0.49,0.74}
\usepackage[pagebackref,breaklinks,colorlinks,allcolors=cvprblue]{hyperref}

\def\paperID{*****} 
\def\confName{CVPR}
\def\confYear{2027}

\title{Is It the Agent or the Retry?\\Separating Selection from Agency in Text-to-Image Generation}

\author{
Jian Xu$^{1,2}$ \qquad Delu Zeng$^{3}$ \qquad Qibin Zhao$^{2}$\\
$^{1}$RIKEN iTHEMS \qquad $^{2}$RIKEN AIP \qquad $^{3}$South China University of Technology\\
{\tt\small jian.xu@riken.jp, dlzeng@scut.edu.cn, qibin.zhao@riken.jp}
}

\begin{document}
\maketitle
\input{sec/0_abstract}
\input{sec/1_intro}
\input{sec/2_related}
\input{sec/3_protocol}
\input{sec/4_experiments}
\input{sec/5_analysis}
\input{sec/6_discussion}
\input{sec/7_conclusion}
{
    \small
    \bibliographystyle{ieeenat_fullname}
    \bibliography{main}
}

\input{sec/X_suppl}

\end{document}

%% file: preamble.tex
\usepackage{booktabs}
\usepackage{multirow}
\usepackage{amsmath,amssymb}
\usepackage{xspace}
\newcommand{\ours}{\textsc{SRR}\xspace}
\DeclareMathOperator*{\argmax}{arg\,max}
\usepackage{amsthm}
\newtheorem{theorem}{Theorem}
\newtheorem{proposition}{Proposition}

%% file: sec/0_abstract.tex
\begin{abstract}
Agentic text-to-image systems wrap generators in loops of planning, routing, verification and refinement, and report large gains on compositional benchmarks, almost always relative to a single sample.
We ask how much of the gain survives against Single-tool Rejection Retry (\ours): the strongest generator already in the system's pool, an off-the-shelf verifier, and retries that stop at the first accepted image under the system's own budget.
Auditing four recent systems with official evaluators, we find that on GenEval \ours matches the accuracy of OctoT2I (0.962 vs.\ 0.96), recovers 63 to 90\% of the gain of GEMS depending on the verifier, and exceeds GenPilot by 12 points with 2.3 generations per image; the effect persists with generators not tuned to the benchmark and on T2I-CompBench++.
A closed-form analysis predicts the accuracy of \ours at every budget and shows that its gain is capped by the verifier's likelihood ratio.
The same analysis also exposes genuine agentic gains: on GenEval 2, even the best of five samples chosen by the evaluator stays 15 points below the score GEMS reports (12 at the upper end of its 95\% interval).
We prove a ceiling on the accuracy that routing across generators can add, bracket it between one and three points on OctoT2I's own pool, and show that what OctoT2I's routing buys is compute (about $2.5\times$ less than \ours at equal accuracy), not accuracy. In our offline distillation experiments selected samples teach the generator no more than random ones, which a short information argument explains.
We end with a reporting checklist for agentic generation papers.
\end{abstract}

%% file: sec/1_intro.tex
\section{Introduction}
\label{sec:intro}

\begin{figure}[t]
\centering
\includegraphics[width=\columnwidth]{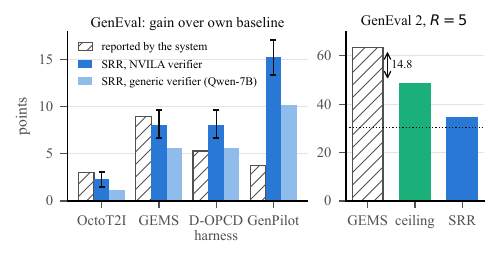}
\caption{\textbf{Selection versus agency.} Left: on GenEval, single-tool rejection retry (\ours) with the strongest tool of each audited system, compared with the gain that system reports over its own single-sample baseline (hatched). Dark blue: NVILA verifier (95\% bootstrap intervals); light blue: generic Qwen2.5-VL-7B verifier with a threshold calibrated on disjoint prompts. Right: on GenEval 2 the picture reverses. Even the best of five samples chosen by the evaluator itself (green) stays 14.8 GM points below GEMS, a lower bound, up to cross-paper protocol differences, on the gain due to planning and prompt rewriting. The dotted line is our single-sample score.}
\label{fig:teaser}
\end{figure}

Text-to-image (T2I) generation is increasingly delivered through agents rather than through a single forward pass.
A typical agentic system receives a request, chooses among several generators or prompts, renders an image, asks a vision-language model (VLM) whether the image satisfies the request, and repeats until it is satisfied or a budget runs out.
Recent systems differ in what the loop optimizes: OctoT2I routes each prompt to the cheapest tool that is predicted to pass a quality threshold~\cite{jiang2026octot2i}; GEMS adds memory and skills on top of a planner and verifier~\cite{he2026gems}; D-OPCD evolves a skill library for a harness and then distills it into the generator~\cite{wang2026dopcd}; GenPilot searches over rewritten prompts with a multi-agent critic~\cite{ye2025genpilot}.
All of them report sizable gains on GenEval~\cite{ghosh2023geneval} and T2I-CompBench++~\cite{huang2025t2icompbench}.

The reference point for these gains is almost always a single image from a single generator.
That comparison conflates two very different sources of improvement.
The first is \emph{selection}: an agent that renders several images and keeps one that a verifier accepts is performing rejection sampling, and rejection sampling improves any benchmark whose metric correlates with the verifier.
The second is everything the agent adds beyond selection: deciding which tool to call, how to rewrite the prompt, which skill to apply, and when to stop.
Only the second is specific to the agent.
The first is available to anyone with a generator, a verifier, and the same number of generations.

This paper measures the two separately.
We define a deliberately simple baseline, \textbf{Single-tool Rejection Retry} (\ours): take the strongest generator that is already in the audited system's tool pool, draw images from it one at a time, score each with a verifier, and return the first image whose score exceeds a threshold (or the best one when the budget is exhausted).
\ours uses no planning, no routing, no prompt rewriting, no memory and no training.
It matches the audited system on the two quantities that the system itself controls, the generation budget and, whenever the system names one, the verifier.

We audit four systems published in 2025 and 2026, using the official GenEval and T2I-CompBench++ evaluators throughout.
Our findings are:
\begin{itemize}
\item \textbf{Selection explains most reported gains.}
On GenEval, \ours matches the accuracy of OctoT2I (0.962 vs.\ 0.96), recovers 63 to 90\% of the gain of GEMS depending on the verifier, and exceeds GenPilot by 12 points with 2.3 generations per image; it does so with tools not tuned to the benchmark and on T2I-CompBench++ (\cref{sec:main,sec:clean}).
A closed form predicts these gains from single-attempt statistics and shows that they are capped by the verifier's likelihood ratio (\cref{prop:srr,sec:curves}).
\item \textbf{The audit also exposes genuine agentic gains.}
On GenEval 2~\cite{kamath2025geneval2}, even picking the best of five samples with the evaluator itself stays 14.8 points below GEMS (12.0 at the upper end of the 95\% interval), a lower bound, up to cross-paper protocol differences, on what its planning and prompt rewriting contribute (\cref{sec:ge2}).
\item \textbf{Routing buys compute, not accuracy; offline distillation adds little.}
We prove a ceiling on the accuracy any routing policy can add over the best single tool (\cref{thm:route}) and bracket it between 0.7 and 2.9 points on OctoT2I's own pool; OctoT2I's routing instead saves about $2.5\times$ compute over \ours, which no training-free cascade we tested matches (\cref{sec:routing}). Offline fine-tuning on selected samples does no better than on random ones, consistent with \cref{prop:info}: accepted samples carry only $-\log\alpha$ nats where acceptance $\alpha$ is high and are missing where it is low (\cref{sec:internal}).
\end{itemize}
We do not claim that agents never help.
Systems that train the generator to reflect on its own failures beat best-of-$N$ at equal budget~\cite{zhuo2025reflectionflow,li2025reflectdit}, D-OPCD's later stages exceed \ours, OctoT2I reaches its accuracy at lower compute, and it retains an unexplained advantage on complex T2I-CompBench++ prompts (\cref{sec:discussion}).
Our point is that the comparison against \ours is cheap to run, changes the conclusions of several recent papers, and should be reported by default.

%% file: sec/2_related.tex
\section{Related Work}
\label{sec:related}

\paragraph{Agentic text-to-image generation.}
LLM and VLM controllers have been placed around diffusion models to plan layouts~\cite{yang2024rpg}, refine prompts from visual feedback~\cite{yang2023idea2img,ye2025genpilot}, call editing and generation tools~\cite{wang2024genartist}, route requests among generators~\cite{jiang2026octot2i}, and accumulate memory and skills across tasks~\cite{he2026gems,wang2026dopcd}.
Most of these systems report improvements over single-sample generation from their backbone; few report a comparison against repeated sampling from that backbone under the same verifier and budget.
GEMS includes a best-of-5 search baseline~\cite{he2026gems}, and we compare against it directly.

\paragraph{Inference-time scaling and verifiers.}
Searching over initial noise with a verifier is an effective way to spend extra compute on diffusion models~\cite{ma2025inference}, and verifier-guided best-of-$N$ is the standard baseline in work that trains generators to reflect on and correct their own outputs~\cite{zhuo2025reflectionflow,li2025reflectdit}.
These papers establish that best-of-$N$ is strong; our contribution is to show that many agentic papers omit it, and to quantify how much of their reported gains it explains.

\paragraph{Internalizing selections.}
Rejection-sampling fine-tuning~\cite{dong2023raft}, best-of-$N$ distillation~\cite{sessa2024bond}, preference optimization~\cite{wallace2024diffusiondpo}, and online reinforcement learning~\cite{black2024ddpo,liu2025flowgrpo,zheng2025diffusionnft} all aim to move a generator toward the outputs a reward or verifier prefers.
For seed selection specifically, prior work learns noise transformations from good and bad seeds~\cite{zhou2024golden,eyring2025hypernoise} or fine-tunes the generator on images from reliable seeds~\cite{li2024seeds}.
D-OPCD distills the prompt-mediated part of an agent's experience into the generator and explicitly leaves verification and sample selection out~\cite{wang2026dopcd}; we test whether that part can be distilled offline at all.

\paragraph{Benchmarks.}
GenEval~\cite{ghosh2023geneval} scores object presence, counts, colors, and positions with a Mask2Former detector~\cite{cheng2022mask2former} and CLIP; T2I-CompBench++~\cite{huang2025t2icompbench} uses BLIP-VQA~\cite{li2022blip}, a detector, depth, and CLIP across eight categories.
Both reward the attributes a VLM verifier checks, and some verifiers used by agents were trained for GenEval~\cite{zhuo2025reflectionflow}.

%% file: sec/3_protocol.tex
\section{Auditing Protocol}
\label{sec:protocol}

\paragraph{Agentic systems.}
We describe an agentic T2I system by a tool pool $\mathcal{T}$ of generators, a verifier $v(x,c)\in[0,1]$ that scores how well image $x$ satisfies request $c$, a maximum number of generations $R$, and a policy that decides at each round which tool to call with which prompt.
The policy may also rewrite the prompt, consult memory or skills, and stop early when $v$ exceeds a threshold $\theta$.

\paragraph{Single-tool Rejection Retry.}
\ours removes the policy and keeps the parts that are not specific to the agent.
Given the strongest tool $t^\star\in\mathcal{T}$, the verifier $v$, the threshold $\theta$, and the budget $R$, \ours draws $x_1,x_2,\dots$ independently from $p_{t^\star}(\cdot\mid c)$ using the original request $c$, returns the first $x_a$ with $v(x_a,c)\ge\theta$, and otherwise returns $\argmax_{a\le R} v(x_a,c)$.
This is exactly the stop rule OctoT2I uses after each routed generation~\cite{jiang2026octot2i}, applied to a single tool.

\paragraph{Choosing the tool, the verifier and the budget.}
We take $t^\star$ to be the strongest single generator in the audited system's own pool according to the audited paper's own tables, so the choice is fixed before we run anything.
When the system uses an open verifier we use the same one (NVILA-Lite-2B-Verifier~\cite{liu2024nvila} for OctoT2I).
Otherwise we use NVILA-Lite-2B-Verifier as an off-the-shelf choice; it is orders of magnitude smaller than the controllers of the audited systems (GPT-5.6, Kimi K2.5, and a 72B Qwen VLM), and we ablate it against Qwen2.5-VL-7B~\cite{bai2025qwen25vl} and an oracle in \cref{sec:curves}.
We use $\theta=0.8$ as in OctoT2I and the score $v=p(\text{yes})/(p(\text{yes})+p(\text{no}))$ at the first answer token, with the verifier's own yes/no prompt.
We set $R$ to the system's maximum number of generations (4 for OctoT2I, 5 for GEMS and D-OPCD) and to 5 for GenPilot, whose budget is far larger.
We report the mean number of generations per evaluated image.

\paragraph{Evaluation.}
We use the official GenEval code (Mask2Former and CLIP, 553 prompts, 4 images per prompt) and the official T2I-CompBench++ evaluators (BLIP-VQA, UniDet, depth, CLIP and 3-in-1; 300 validation prompts per category).
Each evaluated image is a \emph{slot} with its own seeds; attempt $a$ of slot $j$ uses seed $j+Ka$ with $K$ slots per prompt, so single-sample numbers are the first attempts of the same slots.
On T2I-CompBench++ we use two slots per prompt.
OctoT2I's non-spatial column is not on the CLIPScore scale of the official evaluator, so we exclude that category from comparisons and report a seven-category average.
The official 3-in-1 script hard-codes ten images per prompt; we re-implemented its weighting for two.
Absolute single-sample numbers can differ slightly from those reported by the audited papers because of seeds, images per prompt, and prompt subsets, so we report both absolute scores and gains over each paper's own baseline.

\subsection{What selection can buy, and what routing can add}
\label{sec:theory}

GenEval and most T2I-CompBench++ categories score each image as correct or not with respect to a request $c$.
Write $m(x,c)\in\{0,1\}$ for this label, $q_t(c)=\Pr_{x\sim p_t}[m=1]$ for the single-sample accuracy of tool $t$, and describe the verifier on tool $t^\star$ by its per-request true and false acceptance rates $\tau_1(c)=\Pr[v\ge\theta\mid m=1]$ and $\tau_0(c)=\Pr[v\ge\theta\mid m=0]$.
Attempts are independent given $c$ because each uses a fresh seed.

\begin{proposition}[Accuracy and cost of \ours]
\label{prop:srr}
Let $a=q\tau_1$, $b=(1-q)\tau_0$ and $s=a+b$ be the probabilities that one attempt is accepted and correct, accepted and wrong, and accepted.
Then
\begin{align}
\Pr[m(x_{\ours})=1] &= \tfrac{a}{s}\big(1-(1-s)^R\big)+(1-s)^R\,\beta_R, \label{eq:srr}\\
\mathbb{E}[N] &= \big(1-(1-s)^R\big)/s, \nonumber
\end{align}
where $\beta_R$ is the accuracy of the highest-scoring of $R$ rejected attempts.
As $R\to\infty$ the posterior odds of correctness equal the prior odds $q/(1-q)$ times the verifier's likelihood ratio $\tau_1/\tau_0$.
\end{proposition}

Three consequences matter for auditing.
First, \ours is truncated rejection sampling, and its gain is capped by the verifier's likelihood ratio rather than by any property of the agent: a verifier that checks the same attributes as the benchmark (large $\tau_1/\tau_0$) makes the gain large.
Second, cost adapts to difficulty: when most requests are easy, $s$ is close to $q$ and $\mathbb{E}[N]$ is close to one.
Third, \cref{eq:srr} predicts the entire budget curve from per-request statistics of single attempts, which we verify in \cref{sec:curves}.

Routing systems can call a different tool on each attempt and condition that choice on everything observed so far.
The next result bounds what this can add, for any policy and any final selection rule.

\begin{theorem}[Routing ceiling]
\label{thm:route}
Let a system draw $R$ images for request $c$ from tools $t_1,\dots,t_R\in\mathcal{T}$, where each $t_a$ may depend on $c$ and on all previous images and verifier outputs, and let it return one of the $R$ images.
Then its accuracy on $c$ is at most $1-(1-q_{\max}(c))^R$ with $q_{\max}(c)=\max_{t\in\mathcal{T}}q_t(c)$.
Consequently, the accuracy that routing can add over the best single tool $t^\star$ with an oracle verifier is at most
\begin{equation}
H_R=\mathbb{E}_c\big[(1-q_{t^\star}(c))^R-(1-q_{\max}(c))^R\big].
\label{eq:headroom}
\end{equation}
\end{theorem}

The proof conditions on the history: whatever tool is chosen, the next image is wrong with probability at least $1-q_{\max}(c)$, so all $R$ images are wrong with probability at least $(1-q_{\max}(c))^R$, and a returned image can be correct only if some image is (Supp.~\ref{supp:proofs}).
The bound holds for adaptive and cost-aware policies such as OctoT2I's, and it depends only on single-sample accuracies, which any routing paper already has.
Estimating it requires care: replacing $q_t(c)$ by an empirical mean over a few seeds inflates $q_{\max}$, because the maximum of noisy estimates is biased upward.
We therefore estimate the value of per-request routing by cross-fitting, choosing the tool on some seeds and evaluating it on others (\cref{sec:routing}).

Finally, consider distilling an agent's selections into the generator by fine-tuning on accepted images.

\begin{proposition}[Information in offline selections]
\label{prop:info}
Suppose acceptance is a deterministic function of the image and let $\alpha(c)$ be the acceptance rate.
The distribution of accepted images $p_{\mathcal{A}}(\cdot\mid c)$ satisfies $\mathrm{KL}(p_{\mathcal{A}}\,\|\,p)=-\log\alpha(c)$.
If each training request receives $K$ attempts and only requests with at least one acceptance contribute data, the expected tilt carried by a request is $I_K(\alpha)=\big(1-(1-\alpha)^K\big)(-\log\alpha)$, which vanishes both as $\alpha\to1$ and as $\alpha\to0$, where $I_K(\alpha)\approx K\alpha\log(1/\alpha)$.
\end{proposition}

Easy requests are present in the data but differ little from the base model, while the hard requests that selection would change most rarely produce an accepted image at all.
\Cref{prop:info} describes the data, not every way of using it: it does not rule out methods that obtain accepted samples for hard requests by other means, such as more attempts, prompt rewriting or on-policy sampling.
\Cref{sec:internal} measures it on our distillation data, where it is consistent with the failure of the offline protocols we test.

%% file: sec/4_experiments.tex
\section{Auditing Four Agentic Systems}
\label{sec:experiments}

\subsection{GenEval}
\label{sec:main}

\Cref{tab:main} compares each system, as reported in its paper, with \ours built from that system's strongest tool.

\begin{table*}[t]
\centering
\small
\setlength{\tabcolsep}{4.5pt}
\resizebox{\textwidth}{!}{%
\begin{tabular}{lllc cc ccc}
\toprule
& & & & \multicolumn{2}{c}{Reported by the audited paper} & \multicolumn{3}{c}{Ours (official GenEval)} \\
\cmidrule(lr){5-6}\cmidrule(lr){7-9}
System & Strongest tool $t^\star$ & Verifier for \ours & $R$ & Tool alone & System & Tool alone & \ours & Gens / image \\
\midrule
OctoT2I~\cite{jiang2026octot2i} & Flow-GRPO & NVILA-2B (same) & 4 & 0.93 & 0.96 & 0.939 & \textbf{0.962} & 1.26 \\
GEMS~\cite{he2026gems} & Z-Image-Turbo & NVILA-2B & 5 & 0.77 & \textbf{0.86} & 0.746 & 0.827 & 1.99 \\
\quad GEMS Search (best-of-5) & Z-Image-Turbo & MLLM & 5 & 0.77 & 0.81 & & & \\
D-OPCD$^\dagger$~\cite{wang2026dopcd}, skill-free harness & Z-Image-Turbo & NVILA-2B & 5 & 0.782 & 0.835 & 0.746 & 0.827 & 1.99 \\
\quad + evolved skills / + distillation and skills & & & & 0.782 & 0.876 / \textbf{0.918} & & & \\
GenPilot~\cite{ye2025genpilot} & FLUX.1-schnell & NVILA-2B & 5$^\ddagger$ & 0.658 & 0.696 & 0.665 & \textbf{0.818} & 2.30 \\
\midrule
No system (control) & SD3.5-Medium & NVILA-2B & 4 & & & 0.678 & 0.799 & 2.11 \\
\bottomrule
\end{tabular}}
\caption{\textbf{GenEval audit.} \ours uses the strongest generator already in each system's tool pool, the original prompt, and a verifier with OctoT2I's stop rule ($\theta=0.8$), under the system's maximum number of generations $R$. ``Gens / image'' is the mean number of generations actually used. $^\dagger$D-OPCD evaluates one image per prompt on a 194-prompt subset of GenEval with GPT-5.6 as in-loop verifier; its numbers are converted to the 0--1 scale. $^\ddagger$GenPilot renders up to 20 candidate prompts in each of up to 10 refinement cycles; we cap \ours at 5 generations.}
\label{tab:main}
\end{table*}

\paragraph{OctoT2I.}
With the same verifier, threshold and round limit, retrying Flow-GRPO alone gives 0.962 (95\% CI 0.948 to 0.974) against 0.96 for OctoT2I, which routes among five generators with a learned policy; our single-sample Flow-GRPO score (0.939) agrees with the 0.93 OctoT2I reports, and \ours needs 1.26 generations per image because most prompts pass on the first attempt.
This is a match in accuracy at equal maximum budget, not in compute: OctoT2I routes easy prompts to cheaper tools, and \cref{sec:routing} shows that this saves about $2.5\times$ compute.
Per category (Supp.~\cref{tab:geneval_cat,fig:supp_geneval}), \ours is higher than OctoT2I on counting and color attribution, on par on position, and lower on colors.

\paragraph{GEMS and D-OPCD.}
Both systems build on Z-Image-Turbo.
\ours raises it from 0.746 to 0.827 with two generations per image, a gain of 0.081 against the 0.09 GEMS reports (0.77 to 0.86) under the same cap of five iterations, and above GEMS's own best-of-5 search (0.81), which uses the system's large MLLM as verifier.
For D-OPCD, the skill-free harness, which already includes GPT-5.6 planning, prompt refinement and verification, gains 0.053 over direct generation, less than the 0.081 of \ours.
D-OPCD's evolved skills (0.876) and its distilled generator combined with re-evolved skills (0.918) go beyond \ours; we return to this in \cref{sec:discussion}.
D-OPCD evaluates a different prompt subset with one image per prompt, so we compare gains rather than absolute scores.

\paragraph{GenPilot.}
Our single-sample FLUX.1-schnell score (0.665) matches GenPilot's (0.658); its prompt search reaches 0.696, while \ours reaches 0.818 (CI 0.797 to 0.838) with 2.3 generations, four times the gain, without changing the prompt.

\subsection{T2I-CompBench++}
\label{sec:t2ic}

OctoT2I is the only audited system that reports T2I-CompBench++ with an identifiable open tool pool.
\Cref{tab:t2ic} (per category in Supp.~\cref{fig:supp_t2ic}) shows that \ours with Flow-GRPO matches or exceeds OctoT2I on five of the seven comparable categories and on the seven-category average (0.621 vs.\ 0.619).
The verifier is stricter on these longer prompts, so \ours uses 3.3 generations per image.
Two caveats apply.
Our single-sample Flow-GRPO average is 0.012 above the value OctoT2I reports for the same model, so measured against each paper's own baseline \ours recovers about 60\% of OctoT2I's gain (0.015 vs.\ 0.026).
The remaining gap is concentrated in the complex category (0.393 vs.\ 0.430) and, to a lesser extent, numeracy.

\begin{table*}[t]
\centering
\small
\setlength{\tabcolsep}{5pt}
\resizebox{\textwidth}{!}{%
\begin{tabular}{llccccccccc}
\toprule
Tool & Method & Color & Shape & Texture & 2D-Spat. & 3D-Spat. & Numeracy & Complex & Avg (7) & Gens / image \\
\midrule
\multirow{2}{*}{Flow-GRPO$^\ast$} & single sample & 0.812 & 0.614 & 0.719 & 0.520 & 0.437 & 0.675 & 0.374 & 0.593 & 1 \\
 & OctoT2I (5 tools) & 0.834 & 0.627 & 0.732 & 0.527 & 0.433 & \textbf{0.751} & \textbf{0.430} & 0.619 & n/r \\
\midrule
\multirow{3}{*}{Flow-GRPO} & single sample & 0.831 & 0.616 & 0.725 & 0.533 & \textbf{0.445} & 0.706 & 0.384 & 0.606 & 1 \\
 & \ours & \textbf{0.843} & \textbf{0.629} & \textbf{0.752} & \textbf{0.542} & 0.438 & 0.748 & 0.393 & \textbf{0.621} & 3.33 \\
 & \ours, alternate with SANA-1.5 & \textbf{0.843} & \textbf{0.629} & 0.743 & 0.528 & 0.410 & 0.731 & 0.396 & 0.612 & 3.35 \\
\midrule
\multirow{2}{*}{SD3.5-M} & single sample & 0.790 & 0.571 & 0.734 & 0.290 & 0.366 & 0.598 & 0.381 & 0.533 & 1 \\
 & \ours & 0.827 & 0.625 & 0.759 & 0.375 & 0.365 & 0.649 & 0.389 & 0.570 & -- \\
\bottomrule
\end{tabular}}
\caption{\textbf{T2I-CompBench++ audit} with the official evaluators (non-spatial excluded, see \cref{sec:protocol}). $^\ast$As reported by OctoT2I. n/r: not reported. The alternating variant retries Flow-GRPO and SANA-1.5 in turn, with no routing policy.}
\label{tab:t2ic}
\end{table*}

\subsection{Where selection is not enough: GenEval 2}
\label{sec:ge2}

\begin{figure}[t]
\centering
\includegraphics[width=\columnwidth]{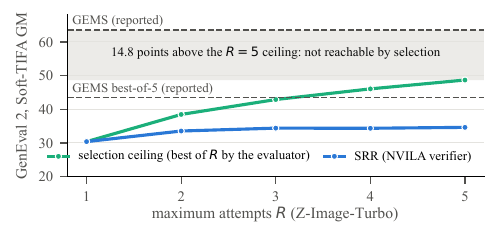}
\caption{\textbf{GenEval 2} (Soft-TIFA GM, official code) for Z-Image-Turbo. \ours saturates near 35 because the verifier rarely accepts long prompts. No rule that selects among $R$ samples can exceed the green ceiling; GEMS (reported) exceeds it by 14.8 points at $R{=}5$. Full numbers, including AM and D-OPCD, are in Supp.~\cref{tab:ge2}.}
\label{fig:ge2}
\end{figure}

GenEval 2~\cite{kamath2025geneval2} has 800 prompts with three to ten atoms each and is scored by Soft-TIFA with Qwen3-VL-8B, one image per prompt; GEMS and D-OPCD both report it for Z-Image-Turbo.
\Cref{fig:teaser} (right), \cref{fig:ge2} and Supp.~\cref{tab:ge2} show a different picture from GenEval.
Our single-sample score (GM 30.4) matches GEMS's baseline (31.0), but \ours adds only 4.2 GM points: the GenEval-trained verifier rarely accepts these long prompts and \ours exhausts its five attempts on most of them.
More importantly, the same argument as \cref{thm:route} bounds \emph{any} selection rule: a system that only chooses among five samples of the tool for the original prompt cannot beat choosing the best of them by the evaluator itself, which reaches 48.7 GM.
GEMS reports 63.5, 14.8 points above the ceiling; D-OPCD's skill-free harness (87.5 AM on its subset) likewise exceeds it (85.2 AM).
The gap is not a sampling artifact: at the upper end of the ceiling's 95\% prompt-bootstrap interval [45.8, 51.5] it is still 12.0 points, and comparing gains over each paper's own baseline (GEMS $+32.5$, ceiling $+18.3$, interval [16.1, 20.5]) also leaves at least 12.0.
It remains a comparison across papers with the same prompts and evaluator but possibly different seeds and sampling settings, as we could not rerun GEMS; we read it as strong evidence, not a certificate, that at least about 12 points of GEMS's gain come from mechanisms other than selection, such as planning and prompt rewriting.
The audit thus cuts both ways: it removes most of the reported gains on GenEval and T2I-CompBench++, and it exposes a large, genuinely agentic gain on GenEval 2.

\subsection{Is the effect an artifact of a benchmark-tuned tool?}
\label{sec:clean}

Flow-GRPO was trained with a GenEval-style reward, but Z-Image-Turbo and FLUX.1-schnell were not, and the base SD3.5-Medium from which Flow-GRPO was fine-tuned gains just as much: \ours lifts it from 0.678 to 0.799 on GenEval and from 0.533 to 0.570 on T2I-CompBench++, most on 2D spatial relations (0.290 to 0.375).

\subsection{Budget curves and verifier ablation}
\label{sec:curves}

\begin{figure*}[t]
\centering
\includegraphics[width=\textwidth]{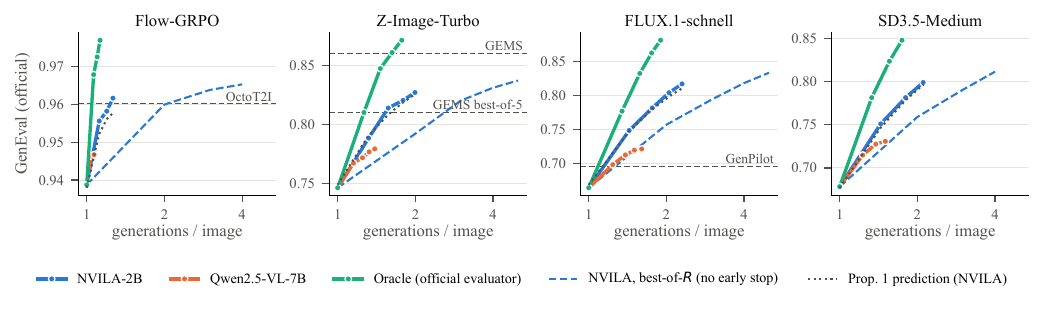}
\caption{\textbf{GenEval against the generations actually spent}, for \ours with each tool and three verifiers ($R=1,\dots,5$; log scale). Dashed gray lines mark the scores reported by the audited systems that use each tool. The dotted line is the prediction of \cref{prop:srr} from per-prompt single-attempt statistics, which overlaps the measured NVILA curve. At equal generations, stopping at the first accepted image (solid) dominates best-of-$R$ without early stopping (dashed blue).}
\label{fig:curves}
\end{figure*}

We generated and scored all attempts of every slot (39,816 images), so any budget and verifier can be evaluated by lookup (\cref{fig:curves}).

\paragraph{\Cref{prop:srr} predicts the curves.}
Estimating $q$, $\tau_1$, $\tau_0$ and the score order of rejected attempts per prompt from the pooled attempts and plugging them into \cref{eq:srr} reproduces the measured accuracy within 0.007 and the mean number of generations within 0.02 at every budget, for all four tools and both verifiers.
Nothing beyond i.i.d.\ rejection sampling is needed to explain \ours. Most of the gain arrives within two or three attempts, and at equal generations stopping at the first acceptance beats keeping the best of $R$, because easy prompts stop early.

\paragraph{The verifier matters, as \cref{prop:srr} says it should.}
At $\theta=0.8$, NVILA-2B accepts 81--93\% of correct images and 5--15\% of wrong ones (likelihood ratio 6 to 18, AUC 0.965 to 0.969 across tools), whereas Qwen2.5-VL-7B accepts 93--98\% of correct and 40--70\% of wrong images (likelihood ratio 1.4 to 2.4, AUC 0.73 to 0.84).
With Qwen, \ours still improves every tool but by much less; even with its threshold calibrated on prompts disjoint from GenEval (Supp.~\cref{tab:calib,fig:supp_calib}) it reaches 0.950, 0.803, 0.766 and 0.776, 0.02 to 0.07 below NVILA, so the gap reflects separation of correct from wrong images, not a mis-set threshold.
This qualifies the audit; for OctoT2I the comparison is exact, since both sides use NVILA.
GenPilot is exceeded with either verifier (calibrated Qwen 0.766 vs.\ 0.696).
GEMS and D-OPCD use general-purpose MLLMs as in-loop verifiers; with a calibrated generic verifier, \ours on Z-Image-Turbo reaches the level of GEMS's own best-of-5 search (0.803 vs.\ 0.81) and a gain (+0.057) comparable to D-OPCD's skill-free harness (+0.053), but not GEMS itself (0.86).
GEMS's gain of 0.09 thus splits into 0.057 (63\%) available to selection with a generic verifier, a further 0.024 from a verifier trained for GenEval-style prompts~\cite{zhuo2025reflectionflow}, and 0.009 to 0.033 from GEMS's planning, memory and skills, depending on which verifier one deems matched; we therefore report 63 to 90\% rather than a single number.
Selection does not trade image quality for accuracy: PickScore, HPSv2.1 and ImageReward stay equal or rise slightly, while diversity among slots drops by 1 to 5\% (Supp.~\cref{tab:quality,fig:supp_quality}); verifier statistics and bootstrap intervals are in Supp.~\cref{tab:verifier,fig:supp_verifier,tab:ci}.
Either way, the quantity that governs selection gains is the verifier's likelihood ratio on the benchmark's attributes, not the sophistication of the agent around it, and the oracle curves (0.87 to 0.98) show that better verifiers would widen the gap further.

\begin{figure*}[t]
\centering
\includegraphics[width=0.9\textwidth]{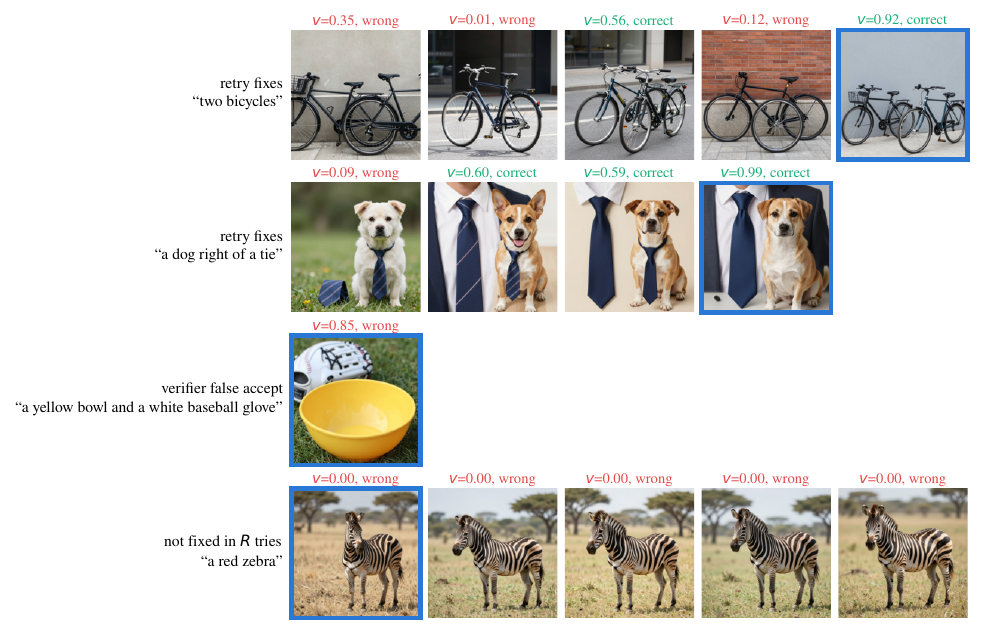}
\caption{\textbf{\ours on Z-Image-Turbo}, GenEval prompts. Each row shows the attempts of one slot with the NVILA score $v$ and the official verdict; the blue frame marks the returned image. Retrying fixes counts and spatial relations that the first sample gets wrong (rows 1--2), even though the strict verifier rejects some correct images (row 1, third attempt). The verifier occasionally accepts a wrong image (row 3). When the generator essentially never produces a correct image (row 4, $q\approx0$), no amount of selection helps; this is where changing the generator or the prompt can pay off.}
\label{fig:qual}
\end{figure*}

%% file: sec/5_analysis.tex
\section{Where Do Agentic Gains Come From?}
\label{sec:analysis}

\Cref{sec:experiments} shows that selection under a verifier explains most of the reported gains.
We now test the two other mechanisms agentic systems rely on: choosing among several generators, and moving what the agent learned into the generator's weights.

\subsection{Routing across generators}
\label{sec:routing}

\begin{figure}[t]
\centering
\includegraphics[width=\columnwidth]{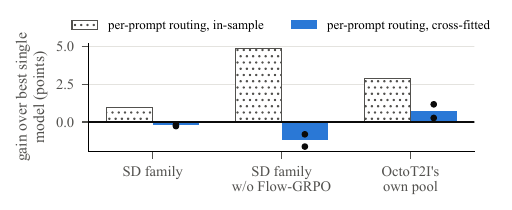}
\caption{\textbf{Routing on GenEval}, one generation per prompt, official evaluator as oracle verifier. Gain of per-prompt routing over the best single model of each pool, chosen in-sample (dotted) or on two seeds and evaluated on the other two (blue; dots are the two folds). The in-sample advantage largely disappears out of sample, as \cref{thm:route} predicts. Full results in Supp.~\cref{tab:routing}.}
\label{fig:routing}
\end{figure}

\Cref{thm:route} bounds what any routing policy can add over the best single tool, and shows that the bound must be estimated out of sample.
We generate four images per GenEval prompt from SD1.5, SDXL, SD3.5-Medium and Flow-GRPO, score all of them with the official evaluator, and compare the best single model with per-prompt routing, both in-sample and cross-fitted (\cref{fig:routing}; Supp.~\cref{tab:routing}), with the official evaluator as an oracle verifier, the setting most favorable to routing.

Per-prompt routing looks useful in-sample, by up to 0.06 when Flow-GRPO is excluded, but chosen on two seeds and evaluated on the other two it is slightly \emph{worse} than the best single model in both pools, the upward bias \cref{thm:route} warns about.
When the best model fails on a prompt, drawing again from it succeeds about twice as often as switching to another model, because a failure says more about the prompt than about the model; likewise, alternating Flow-GRPO with SANA-1.5 lowers the T2I-CompBench++ score (\cref{tab:t2ic}).

\paragraph{OctoT2I's own tool pool.}
The models above are of one family, which may understate complementarity.
We therefore also evaluate the five tools OctoT2I routes among, Flow-GRPO (0.939), SANA-1.5 (0.798), SANA-Sprint (0.738), SDXL-Turbo (0.556) and SD-Turbo (0.480), four images per prompt each.
Here routing has some real value, and three quantities must be kept apart.
The ceiling $H_R$ of \cref{eq:headroom} is a population quantity; its plug-in estimate from four seeds per tool, $\hat H_1=0.029$ and $\hat H_4=0.020$, is biased upward.
The gain achieved by one particular cross-fitted per-prompt router, 0.007 at one generation and 0.006 at two (switching away from Flow-GRPO on 5\% of prompts), is biased downward as an estimate of $H_1$, because the held-out accuracy of whatever tool is chosen is at most $q_{\max}(c)$.
At $R{=}1$ the two therefore bracket the ceiling in expectation, $0.007\lesssim H_1\lesssim0.029$: a better router than ours could add up to about three points of accuracy at one generation, and about two at four.
OctoT2I's reported 0.96 does not exceed the 0.962 of \ours, so in its case routing adds no measurable accuracy.

\begin{figure}[t]
\centering
\includegraphics[width=\columnwidth]{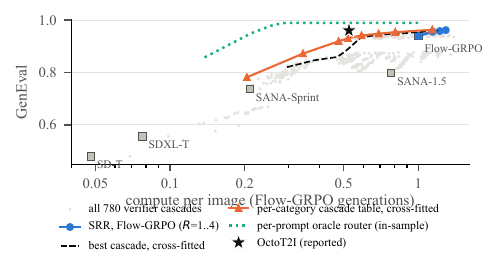}
\caption{\textbf{Accuracy against compute} on OctoT2I's tool pool (GenEval, official evaluator, NVILA at $\theta=0.8$, at most four generations). Compute is measured latency on one A6000 in units of one Flow-GRPO generation, including verifier calls. OctoT2I's point converts its reported time per image (10.02\,s vs.\ 19.07\,s for Flow-GRPO alone) by the same unit. It reaches the accuracy of \ours at about 40\% of the compute, which none of the training-free cascades we tested matches.}
\label{fig:cost}
\end{figure}

\paragraph{What routing buys: compute.}
OctoT2I aims to meet a quality threshold cheaply, which accuracy alone misses.
We measured the latency of every tool in its pool and of the verifier on one A6000, in units of one Flow-GRPO generation (3.98\,s; a verifier call costs 0.02).
\ours reaches 0.962 at 1.29; OctoT2I reports 0.96 at 0.53 in these units, including its LLM controller (\cref{fig:cost}).
No training-free alternative we tried comes close at that cost: the best of all 780 verifier cascades over the pool with up to four generations reaches 0.88 when chosen on half of the prompts and evaluated on the other half, and a per-category lookup table of cascades reaches 0.93, using GenEval's own category labels; both need about 0.8 to reach 0.95.
A per-prompt oracle that knew every outcome in advance would reach 0.99 at 0.29, so the room for cost-aware routing is large, and OctoT2I exploits part of it.
OctoT2I's routing thus buys about $2.5\times$ less compute than \ours at equal accuracy, not higher accuracy; as its time is measured on other hardware, the ratio is approximate.

\subsection{Internalizing selections}
\label{sec:internal}

If selection explains most of an agent's gain, a natural next step is to move it into the generator, as D-OPCD does for prompt refinement~\cite{wang2026dopcd}.
We take SD3.5-Medium at 512px as the generator and 2,000 prompts sampled from the Flow-GRPO GenEval training prompts, which are disjoint from GenEval, with four samples each.
A verifier implementing GenEval's rules accepts or rejects each sample.
We train LoRA students that see only the original prompt on (a) the first accepted sample (\emph{selected SFT}); (b) a random sample from the same prompts (\emph{random SFT}), which removes image-level selection only; (c) accepted versus rejected pairs with Diffusion-DPO~\cite{wallace2024diffusiondpo}; (d) all samples with an accepted or rejected tag; and (e) accepted samples with their initial noise~\cite{li2024seeds}.

Supp.~\cref{tab:distill} gives the full results and Supp.~\cref{fig:supp_prop2} the acceptance distribution. Fine-tuning on self-generated data helps (base 0.678 to 0.729), but not because of selection: selected SFT differs from random SFT by $+0.006\pm0.008$ (0.733 vs.\ 0.729), Diffusion-DPO is worse (0.670 to 0.683), the tagged variant reaches 0.717, and pairing images with their true initial noise reaches 0.742, within two standard errors.
All offline variants stay far below what selection achieves at test time (0.860) and what online RL achieves (0.939).
These conclusions are limited to the protocols we test; D-OPCD's distillation stage, which also changes the prompts, does improve on its harness~\cite{wang2026dopcd}.

\Cref{prop:info} predicts where the information sits, and the teacher pool confirms it (Supp.~\cref{supp:why,fig:supp_prop2}): 19.6\% of the prompts have no accepted sample in four attempts, prompts with acceptance at least 0.75 carry only 0.08 nats of tilt, and those with intermediate acceptance carry 77\% of the best-of-4 gain but are only 27\% of the data.

%% file: sec/6_discussion.tex
\section{Discussion}
\label{sec:discussion}

\paragraph{When agents do beat \ours.}
Results that go beyond \ours share a pattern: the agent changes the generator or the information available to it, rather than only choosing among its samples.
ReflectionFlow reaches 0.91 on GenEval against 0.85 for verifier-guided search at 32 samples~\cite{zhuo2025reflectionflow}, and Reflect-DiT improves SANA-1.0 from 0.75 with best-of-20 to 0.81~\cite{li2025reflectdit}, both by training the generator to revise its outputs.
D-OPCD's evolved skills and distillation stage (0.876, 0.918) also exceed \ours~\cite{wang2026dopcd}, as do GEMS and D-OPCD on GenEval 2 (\cref{sec:ge2}), where long prompts benefit from rewriting and planning.
OctoT2I's lead on complex T2I-CompBench++ prompts (\cref{sec:t2ic}) may come from routing or from protocol differences.

\paragraph{A reporting checklist.}
Papers proposing agentic generation systems should report (i) \ours with the strongest tool, the same verifier and budget, plus a generic verifier if the system's is not open; (ii) accuracy against measured compute, including verifier and controller calls; (iii) whether the verifier was trained on benchmark-like data; (iv) routing choices evaluated on held-out seeds; and (v) prompt subset, images per prompt, seeds and evaluator version.

\paragraph{Limitations.}
Several systems rely on paid APIs or unreleased components, so we audit reported numbers; seed, subset and hardware differences remain, notably for the GenEval 2 gap and OctoT2I's compute ratio. Agents may add more on open-ended requests, aesthetics, or world knowledge.

%% file: sec/7_conclusion.tex
\section{Conclusion}
\label{sec:conclusion}

Agentic T2I systems are usually compared with a single sample, which credits the agent with gains that rejection sampling alone provides.
Matching the tool, the verifier and the budget with \ours, we find that selection explains most of the reported GenEval and T2I-CompBench++ gains of four recent systems, that OctoT2I's routing buys compute, not accuracy, and that offline fine-tuning on selected samples adds nothing over random ones.
Agents do more than select on GenEval 2, where planning and prompt rewriting add at least about 12 points beyond any selection among the tool's samples.
Reporting \ours next to agentic results is cheap and makes such gains attributable.

%% file: sec/X_suppl.tex
\clearpage
\maketitlesupplementary
\setcounter{section}{0}\setcounter{table}{0}\setcounter{figure}{0}\setcounter{equation}{0}
\renewcommand{\thesection}{S\arabic{section}}\renewcommand{\theHsection}{S\arabic{section}}
\renewcommand{\thetable}{S\arabic{table}}\renewcommand{\theHtable}{S\arabic{table}}
\renewcommand{\thefigure}{S\arabic{figure}}\renewcommand{\theHfigure}{S\arabic{figure}}
\renewcommand{\theequation}{S\arabic{equation}}\renewcommand{\theHequation}{S\arabic{equation}}

\section{Proofs}
\label{supp:proofs}

Throughout, the request $c$ is fixed and each attempt uses a fresh seed, so given the tool used at attempt $a$, the image $x_a$ is independent of all earlier images and verifier outputs.

\paragraph{Proposition~\ref{prop:srr}.}
Each \ours attempt draws from the same tool $t^\star$, so attempts are i.i.d.\ and each falls into one of three events: accepted and correct (probability $a=q\tau_1$), accepted and wrong ($b=(1-q)\tau_0$), or rejected ($1-s$).
\ours stops at the first acceptance, hence it returns a correct image at attempt $k\le R$ with probability $(1-s)^{k-1}a$.
Summing over $k$ gives $a\,(1-(1-s)^R)/s$.
With probability $(1-s)^R$ all attempts are rejected and \ours returns the highest-scoring one, which is correct with probability $\beta_R$ by definition.
The number of attempts satisfies $\Pr[N\ge k]=(1-s)^{k-1}$ for $k\le R$, so $\mathbb{E}[N]=\sum_{k=1}^R(1-s)^{k-1}=(1-(1-s)^R)/s$.
As $R\to\infty$ the accuracy tends to $a/s$, whose odds are $a/b=\frac{q}{1-q}\cdot\frac{\tau_1}{\tau_0}$.

\paragraph{Theorem~\ref{thm:route}.}
Let $\mathcal{H}_{a-1}$ collect the images and verifier outputs before attempt $a$, together with any internal randomness of the policy; the tool $t_a$ is $\mathcal{H}_{a-1}$-measurable.
Let $W_a=\{m(x_a,c)=0\}$.
Because $x_a\sim p_{t_a}(\cdot\mid c)$ with a fresh seed, $\Pr[W_a\mid\mathcal{H}_{a-1}]=1-q_{t_a}(c)\ge1-q_{\max}(c)$.
By the tower property, $\Pr[\cap_{a\le R}W_a]=\mathbb{E}\big[\prod_{a\le R}\Pr[W_a\mid\mathcal{H}_{a-1}]\big]\ge(1-q_{\max}(c))^R$.
The returned image is one of $x_1,\dots,x_R$, so it can be correct only on the complement of $\cap_a W_a$, which proves the first claim.
\ours with tool $t^\star$ and an oracle verifier returns a correct image whenever one exists, so its accuracy is exactly $1-(1-q_{t^\star}(c))^R$; subtracting and averaging over $c$ gives \cref{eq:headroom}.
If a system also rewrites prompts, the same argument applies with $\mathcal{T}$ replaced by the set of (generator, prompt) pairs it can choose from.
For the estimation remark: with unbiased estimates $\hat q_t(c)$, convexity of the maximum gives $\mathbb{E}[\max_t\hat q_t(c)]\ge\max_t q_t(c)$, so the plug-in estimate of $H_R$ is biased upward; a tool selected on one set of seeds and evaluated on disjoint seeds yields an unbiased estimate of the accuracy of that selection rule, which cannot exceed the bound.

\paragraph{Proposition~\ref{prop:info}.}
With a deterministic acceptance set $\mathcal{A}_c$, $p_{\mathcal{A}}(x\mid c)=p(x\mid c)\,\mathbf{1}[x\in\mathcal{A}_c]/\alpha(c)$, so $\mathrm{KL}(p_{\mathcal{A}}\|p)=\mathbb{E}_{p_{\mathcal{A}}}[\log(1/\alpha)]=-\log\alpha$.
A request yields at least one accepted image among $K$ attempts with probability $1-(1-\alpha)^K$, and for small $\alpha$ this is $K\alpha+O(\alpha^2)$, giving $I_K(\alpha)\approx K\alpha\log(1/\alpha)\to0$; as $\alpha\to1$, $-\log\alpha\to0$.

\section{Additional results}
\label{supp:results}

\begin{figure*}[t]
\centering
\includegraphics[width=\textwidth]{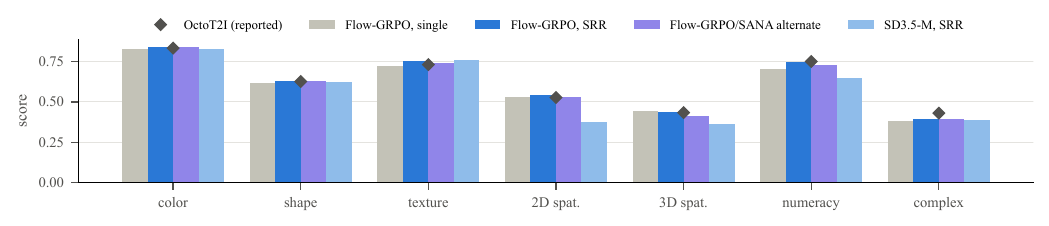}
\caption{\textbf{T2I-CompBench++ per category} (official evaluators). Alternating Flow-GRPO with SANA-1.5 on retry does not help, SD3.5-Medium with \ours lags on spatial categories, and OctoT2I (reported, diamonds) keeps its lead on complex prompts.}
\label{fig:supp_t2ic}
\end{figure*}

\begin{table}[h]
\centering
\small
\setlength{\tabcolsep}{3pt}
\resizebox{\columnwidth}{!}{%
\begin{tabular}{lcccc}
\toprule
& \multicolumn{2}{c}{4 models} & \multicolumn{2}{c}{3 models (no Flow-GRPO)} \\
\cmidrule(lr){2-3}\cmidrule(lr){4-5}
& $K{=}1$ & $K{=}4$ & $K{=}1$ & $K{=}4$ \\
\midrule
Best single model & 0.937 & 0.975 & 0.666 & 0.837 \\
Best static mix of models & 0.937 & 0.975 & 0.666 & 0.837 \\
Per-prompt routing, in-sample & 0.946 & 0.982 & 0.715 & 0.895 \\
\midrule
\multicolumn{5}{l}{\emph{Cross-fitted, $K{=}1$ (choose on two seeds, evaluate on the other two)}} \\
Best single model & \multicolumn{2}{c}{0.931 / 0.942} & \multicolumn{2}{c}{0.667 / 0.665} \\
Per-prompt routing & \multicolumn{2}{c}{0.929 / 0.939} & \multicolumn{2}{c}{0.651 / 0.656} \\
\midrule
\multicolumn{5}{l}{\emph{Success of a second draw after the best model fails}} \\
Same model again & \multicolumn{2}{c}{0.43} & \multicolumn{2}{c}{0.32} \\
Best other model & \multicolumn{2}{c}{0.19} & \multicolumn{2}{c}{0.22} \\
\bottomrule
\end{tabular}}
\caption{\textbf{Routing on GenEval} with the official evaluator as an oracle verifier, $K$ generations per prompt. The in-sample advantage of per-prompt routing disappears when the choice is made on held-out seeds, as \cref{thm:route} predicts.}
\label{tab:routing}
\end{table}

\begin{table}[h]
\centering
\small
\setlength{\tabcolsep}{4pt}
\resizebox{\columnwidth}{!}{%
\begin{tabular}{lccc}
\toprule
Z-Image-Turbo, GenEval 2 & AM & GM & Gens / image \\
\midrule
Single sample (ours) & 77.2 & 30.4 & 1 \\
\ours, NVILA, $\theta=0.8$ & 79.3 & 34.6 & 4.73 \\
\ours, NVILA, $\theta=0.99$ & 79.4 & 34.9 & 4.96 \\
Selection ceiling: best of 5 by the evaluator & 85.2 & 48.7 & 5 \\
\midrule
GEMS$^\ast$, baseline / best-of-5 search / system & & 31.0 / 43.5 / \textbf{63.5} & $\le5$ \\
D-OPCD$^\ast$, baseline / skill-free harness$^\dagger$ & 78.1 / \textbf{87.5} & & $\le5$ \\
\bottomrule
\end{tabular}}
\caption{\textbf{GenEval 2}, Soft-TIFA (official code). $^\ast$Reported. $^\dagger$On a 280-prompt subset. Bootstrap 95\% intervals are about $\pm1.1$ (AM) and $\pm2.7$ (GM). No rule that selects among five samples can exceed the ceiling row; GEMS exceeds it by 14.8 GM points.}
\label{tab:ge2}
\end{table}

\begin{table}[h]
\centering
\small
\setlength{\tabcolsep}{2.6pt}
\resizebox{\columnwidth}{!}{%
\begin{tabular}{lccccccc}
\toprule
& Single & Two & Count & Colors & Pos. & Attr. & Overall \\
\midrule
Flow-GRPO$^\ast$ & 0.99 & 0.99 & 0.88 & 0.95 & 0.95 & 0.86 & 0.93 \\
OctoT2I$^\ast$ & 1.00 & 0.99 & 0.95 & 0.94 & 1.00 & 0.86 & 0.96 \\
\midrule
Flow-GRPO & 1.000 & 0.992 & 0.941 & 0.907 & 0.963 & 0.830 & 0.939 \\
\ours, $R{=}2$ & 1.000 & 0.995 & 0.981 & 0.912 & 0.988 & 0.858 & 0.956 \\
\ours, $R{=}4$ & 1.000 & 0.995 & 0.984 & 0.920 & 0.995 & 0.875 & 0.962 \\
Oracle, $R{=}4$ & 1.000 & 1.000 & 0.997 & 0.947 & 0.998 & 0.920 & 0.977 \\
\bottomrule
\end{tabular}}
\caption{\textbf{GenEval per category, OctoT2I audit.} $^\ast$As reported by OctoT2I. Oracle uses the official evaluator itself as verifier and bounds what any selection rule can achieve with four Flow-GRPO samples.}
\label{tab:geneval_cat}
\end{table}

\begin{figure*}[t]
\centering
\includegraphics[width=\textwidth]{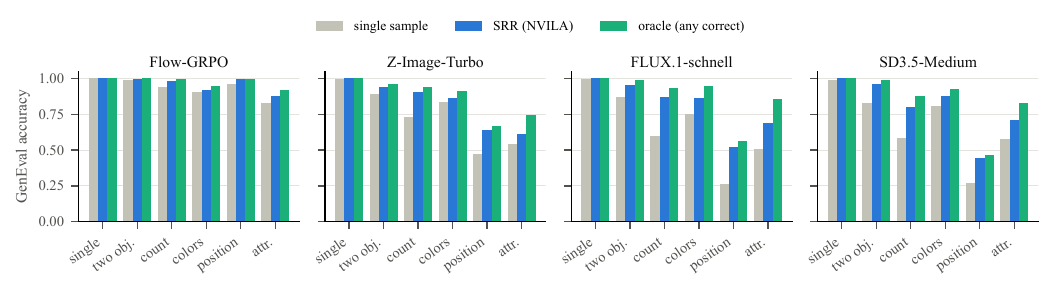}
\caption{\textbf{GenEval per category} for the single sample, \ours (NVILA, $\theta=0.8$) and the oracle that accepts if any of the $R$ attempts is correct. \ours closes most of the gap to the oracle on counting, position and two objects; the largest remaining gap is color attribution (\eg 0.69 vs.\ 0.86 for FLUX.1-schnell).}
\label{fig:supp_geneval}
\end{figure*}

\begin{table}[h]
\centering
\small
\setlength{\tabcolsep}{3pt}
\resizebox{\columnwidth}{!}{%
\begin{tabular}{llccc}
\toprule
Tool ($R$) & Verifier & $\theta=0.8$ & $\theta^\star$ (calibrated) & Gens / image at $\theta^\star$ \\
\midrule
Flow-GRPO (4) & NVILA-2B & 0.962 & 0.965 ($\theta^\star$=0.99) & 1.98 \\
 & Qwen2.5-VL-7B & 0.947 & 0.950 (0.995) & 1.40 \\
Z-Image-Turbo (5) & NVILA-2B & 0.827 & 0.836 (0.99) & 2.85 \\
 & Qwen2.5-VL-7B & 0.779 & 0.803 (0.999) & 4.41 \\
FLUX.1-schnell (5) & NVILA-2B & 0.818 & 0.834 (0.995) & 3.59 \\
 & Qwen2.5-VL-7B & 0.722 & 0.766 (0.999) & 4.86 \\
SD3.5-Medium (4) & NVILA-2B & 0.799 & 0.811 (0.995) & 3.12 \\
 & Qwen2.5-VL-7B & 0.731 & 0.776 (0.999) & 3.66 \\
\bottomrule
\end{tabular}}
\caption{\textbf{Verifier thresholds calibrated on disjoint prompts.} $\theta^\star$ maximizes the simulated accuracy of \ours on 500 Flow-GRPO GenEval-training prompts (4 images each, official-evaluator labels) for each tool and verifier, and is applied unchanged to GenEval. A single threshold chosen on the SD3.5-Medium pool gives the same GenEval scores within 0.004.}
\label{tab:calib}
\end{table}

\begin{figure*}[t]
\centering
\includegraphics[width=\textwidth]{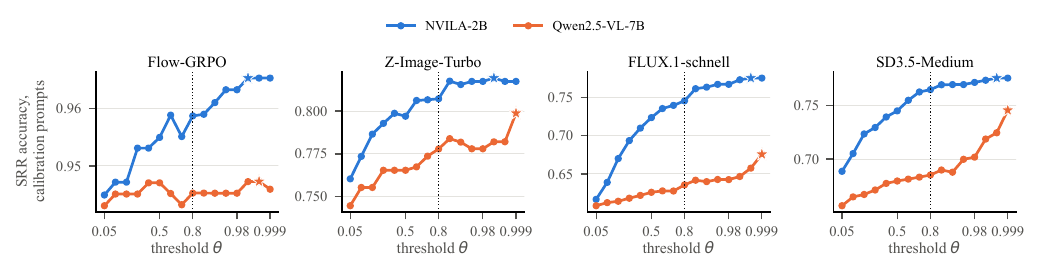}
\caption{\textbf{Threshold calibration} on 500 calibration prompts disjoint from GenEval ($R=4$, one slot). Stars mark the best threshold; the dotted line is the default $\theta=0.8$. Qwen2.5-VL-7B stays below NVILA at every threshold.}
\label{fig:supp_calib}
\end{figure*}

\begin{table}[h]
\centering
\small
\setlength{\tabcolsep}{4pt}
\resizebox{\columnwidth}{!}{%
\begin{tabular}{lcc}
\toprule
Student (SD3.5-M, 512px) & GenEval & vs.\ random SFT \\
\midrule
Base model & 0.678 & $-0.049\pm0.010$ \\
(b) Random SFT (control) & 0.729 & -- \\
(a) Selected SFT & 0.733 & $+0.006\pm0.008$ \\
(c) Diffusion-DPO, three settings & 0.670--0.683 & $\le -0.044$ \\
(d) Accept/reject tag & 0.717 & $-0.011\pm0.009$ \\
(e) Selected + true initial noise & 0.742 & $+0.014\pm0.008$ \\
\midrule
Base, oracle best-of-4 & 0.860 & \\
Flow-GRPO (online RL) & 0.939 & \\
\bottomrule
\end{tabular}}
\caption{\textbf{Offline internalization of selections}, official GenEval. Differences are prompt-level means with standard errors over the 553 prompts. Image-level selection adds nothing measurable over random self-generated data from the same prompts.}
\label{tab:distill}
\end{table}

\begin{figure}[t]
\centering
\includegraphics[width=\columnwidth]{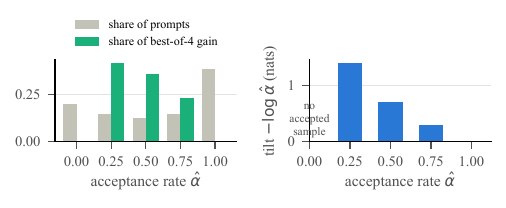}
\caption{\textbf{Why offline selection carries little signal} (\cref{prop:info}). Left: distribution of the per-prompt acceptance rate $\hat\alpha$ of the distillation teacher pool, and each bin's share of the best-of-4 gain. Right: the tilt $-\log\hat\alpha$ that accepted samples carry relative to the base model. Prompts with $\hat\alpha\le0.5$ carry 77\% of the gain, but there accepted samples are scarce, and for the 20\% of prompts with $\hat\alpha=0$ they do not exist; where they are plentiful ($\hat\alpha=0.75$) they differ from the base model by only 0.29 nats.}
\label{fig:supp_prop2}
\end{figure}

\begin{table}[h]
\centering
\small
\setlength{\tabcolsep}{3pt}
\resizebox{\columnwidth}{!}{%
\begin{tabular}{llll}
\toprule
Tool & Single sample & \ours (NVILA, $\theta=0.8$) & Gain \\
\midrule
\multicolumn{4}{l}{\emph{GenEval (official), 95\% intervals}} \\
Flow-GRPO & 0.939 [0.923, 0.954] & 0.962 [0.948, 0.974] & +0.023 [0.015, 0.031] \\
Z-Image-Turbo & 0.746 [0.721, 0.771] & 0.827 [0.803, 0.850] & +0.081 [0.066, 0.096] \\
FLUX.1-schnell & 0.665 [0.643, 0.687] & 0.818 [0.797, 0.838] & +0.153 [0.137, 0.171] \\
SD3.5-Medium & 0.678 [0.655, 0.702] & 0.799 [0.777, 0.820] & +0.120 [0.104, 0.137] \\
\midrule
\multicolumn{4}{l}{\emph{T2I-CompBench++, seven-category average}} \\
Flow-GRPO & 0.606 [0.594, 0.617] & 0.621 [0.609, 0.633] & +0.015 [0.012, 0.019] \\
SD3.5-Medium & 0.533 [0.522, 0.543] & 0.570 [0.559, 0.581] & +0.037 [0.031, 0.044] \\
Flow-GRPO, alternate with SANA-1.5 & & 0.612 [0.600, 0.623] & $-0.009$ [$-0.014$, $-0.005$] vs.\ \ours \\
\bottomrule
\end{tabular}}
\caption{\textbf{Bootstrap intervals} (2,000 resamples of prompts, stratified by task). OctoT2I's reported 0.96 (GenEval) and 0.619 (T2I-CompBench++) lie inside the intervals of \ours with Flow-GRPO; GEMS's 0.86 lies above the interval for Z-Image-Turbo; GenPilot's 0.696 lies far below the interval for FLUX.1-schnell.}
\label{tab:ci}
\end{table}

\begin{table}[h]
\centering
\small
\setlength{\tabcolsep}{3pt}
\resizebox{\columnwidth}{!}{%
\begin{tabular}{lcccc}
\toprule
Tool & $\Delta$PickScore & $\Delta$HPSv2.1 & $\Delta$ImageReward & Diversity (single $\to$ \ours) \\
\midrule
Flow-GRPO & +0.008 [$-0.001$, 0.017] & +0.0003 [0.0000, 0.0006] & +0.010 [0.005, 0.016] & 0.0094 $\to$ 0.0093 \\
Z-Image-Turbo & +0.067 [0.044, 0.091] & +0.0016 [0.0010, 0.0023] & +0.058 [0.039, 0.078] & 0.0107 $\to$ 0.0103 \\
FLUX.1-schnell & +0.121 [0.090, 0.153] & +0.0031 [0.0022, 0.0040] & +0.122 [0.094, 0.152] & 0.0172 $\to$ 0.0166 \\
SD3.5-Medium & +0.095 [0.066, 0.125] & +0.0035 [0.0026, 0.0044] & +0.139 [0.112, 0.169] & 0.0146 $\to$ 0.0139 \\
\bottomrule
\end{tabular}}
\caption{\textbf{Image quality and diversity} of the images returned by \ours versus single samples on GenEval (paired differences with 95\% intervals). Diversity is one minus the mean pairwise cosine similarity of CLIP-H embeddings among the four slots of a prompt. Selection does not trade image quality for compositional accuracy; diversity among slots drops by 3 to 5\% (relative).}
\label{tab:quality}
\end{table}

\begin{figure*}[t]
\centering
\includegraphics[width=\textwidth]{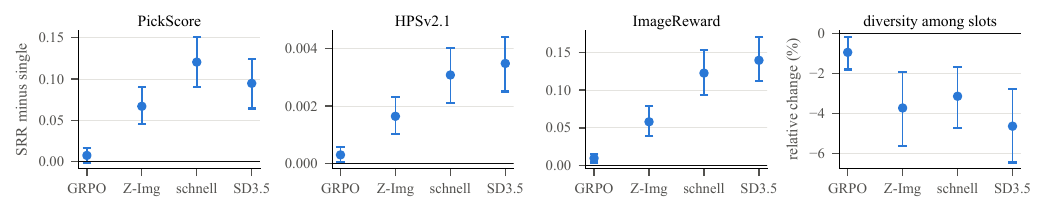}
\caption{\textbf{Quality and diversity}: \ours minus single sample on GenEval, paired 95\% bootstrap intervals. Quality scores do not drop; diversity among the four slots of a prompt falls by 1 to 5\% (relative).}
\label{fig:supp_quality}
\end{figure*}

\begin{table}[h]
\centering
\small
\setlength{\tabcolsep}{3pt}
\resizebox{\columnwidth}{!}{%
\begin{tabular}{lcccccc}
\toprule
& \multicolumn{3}{c}{NVILA-Lite-2B-Verifier} & \multicolumn{3}{c}{Qwen2.5-VL-7B} \\
\cmidrule(lr){2-4}\cmidrule(lr){5-7}
Tool & TPR / FPR & LR & AUC & TPR / FPR & LR & AUC \\
\midrule
Flow-GRPO & 0.93 / 0.15 & 6.2 & 0.965 & 0.98 / 0.70 & 1.4 & 0.734 \\
Z-Image-Turbo & 0.90 / 0.11 & 8.4 & 0.966 & 0.97 / 0.54 & 1.8 & 0.770 \\
FLUX.1-schnell & 0.83 / 0.06 & 13.5 & 0.965 & 0.93 / 0.42 & 2.2 & 0.813 \\
SD3.5-Medium & 0.81 / 0.05 & 18.1 & 0.969 & 0.95 / 0.40 & 2.4 & 0.842 \\
\bottomrule
\end{tabular}}
\caption{\textbf{Verifier statistics} on the GenEval grids at $\theta=0.8$, against official-evaluator labels: true and false acceptance rates ($\tau_1$, $\tau_0$ in \cref{prop:srr}), likelihood ratio $\tau_1/\tau_0$ and AUC.}
\label{tab:verifier}
\end{table}

\begin{figure*}[t]
\centering
\includegraphics[width=\textwidth]{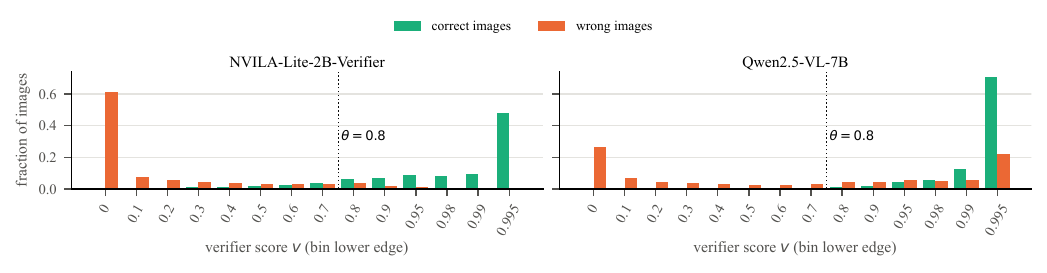}
\caption{\textbf{Verifier score distributions} over all GenEval grid images of the four tools, split by the official evaluator's label. NVILA scores 61\% of wrong images below 0.1 and only 7\% above $\theta$; Qwen2.5-VL-7B scores 46\% of wrong images above $\theta$. This lower likelihood ratio caps its gain, as \cref{prop:srr} predicts.}
\label{fig:supp_verifier}
\end{figure*}

\begin{table}[h]
\centering
\small
\resizebox{\columnwidth}{!}{%
\begin{tabular}{lccc}
\toprule
SD3.5-Medium (512px), GenEval 2 & AM & GM & Gens / image \\
\midrule
Single sample & 64.9 & 19.2 & 1 \\
\ours, NVILA, $\theta=0.8$ & 69.8 & 24.9 & 4.86 \\
Selection ceiling (best of 5) & 76.9 & 34.3 & 5 \\
\bottomrule
\end{tabular}}
\caption{\textbf{GenEval 2 control} with the base SD3.5-Medium.}
\label{tab:ge2sd}
\end{table}

\clearpage
\section{Where the information in offline selections sits}
\label{supp:why}

\Cref{prop:info} predicts where the information sits, and the teacher pool confirms it.
19.6\% of the 2,000 prompts have no accepted sample in four attempts and contribute nothing; they concentrate in position (47\% of those prompts), color attribution and counting, where selection matters most at test time.
Among contributing prompts, 48\% had all four samples accepted, and prompts with acceptance at least 0.75 carry only 0.08 nats of tilt ($-\log\hat\alpha$) on average.
Prompts with intermediate acceptance (0.25 to 0.5) carry 1.07 nats each and 77\% of the best-of-4 gain, but are only 27\% of the data.
Selected and random SFT thus differ on few prompts and neither sees the hard ones, consistent with their indistinguishable scores; online RL avoids both problems by repeatedly sampling every prompt from the current model~\cite{liu2025flowgrpo,zheng2025diffusionnft}.

\section{An editing agent: GenArtist on T2I-CompBench}
\label{supp:genartist}

GenArtist~\cite{wang2024genartist} plans with GPT-4V, calls generation tools and, unlike the other audited systems, \emph{edits} its images through a tree of correction steps.
We run \ours with PixArt-$\alpha$ (PixArt-XL-2-1024-MS), the strongest general-purpose generator in its pool, two slots per prompt, NVILA, $\theta=0.8$, $R=4$ (3.8 generations per image on average), and evaluate with the official T2I-CompBench evaluators (\cref{tab:genartist}).
\ours again adds consistently to its tool (color $+0.09$, shape and texture $+0.08$, 2D spatial $+0.07$).
The comparison with GenArtist itself is inconclusive for three reasons.
First, the PixArt-$\alpha$ row in GenArtist's table is copied from T2I-CompBench++, where it is labeled as a fine-tuned variant; our measurement of the public checkpoint is much lower, and inspection confirms strong attribute leakage across objects that does not depend on numerical precision (fp16, bf16 and fp32 give the same images).
Second, GenArtist's reported scores exceed \ours by a wide margin, consistent with \cref{sec:discussion}: editing changes the image rather than selecting among samples.
Third, an independent re-run of GenArtist reported by OctoT2I~\cite{jiang2026octot2i} obtains 0.609 (color), 0.192 (2D spatial) and 0.329 (complex), below GenArtist's own numbers and below \ours on two of the three categories.

\begin{table}[h]
\centering
\small
\setlength{\tabcolsep}{3pt}
\resizebox{\columnwidth}{!}{%
\begin{tabular}{lcccccc}
\toprule
& Color & Shape & Texture & 2D-Spat. & Non-spat. & Complex \\
\midrule
PixArt-$\alpha$-ft$^\ast$ (as cited by GenArtist) & 0.669 & 0.493 & 0.648 & 0.206 & 0.320 & 0.343 \\
GenArtist$^\ast$ (reported) & 0.848 & 0.695 & 0.771 & 0.544 & 0.335 & 0.450 \\
GenArtist, re-run by OctoT2I$^\ast$ & 0.609 & 0.480 & 0.524 & 0.192 & -- & 0.329 \\
\midrule
PixArt-$\alpha$ public checkpoint, single (ours) & 0.434 & 0.380 & 0.451 & 0.206 & 0.308 & 0.328 \\
PixArt-$\alpha$, \ours (ours) & 0.525 & 0.458 & 0.544 & 0.280 & 0.310 & 0.349 \\
\bottomrule
\end{tabular}}
\caption{\textbf{GenArtist audit on T2I-CompBench}, official evaluators. $^\ast$Reported numbers. Our non-spatial column is the official CLIPScore, on the same scale as GenArtist's.}
\label{tab:genartist}
\end{table}